\documentclass[runningheads]{llncs}
\usepackage{graphicx}

\usepackage{tikz}
\usepackage{comment}
\usepackage{amsmath,amssymb} 
\usepackage{color}
\usepackage{booktabs}
\usepackage{multirow}

\begin{document}
\pagestyle{headings}
\mainmatter

\title{Learning Discriminative Feature with CRF for Unsupervised Video Object Segmentation} 


\titlerunning{DFNet for Unsupervised Video Object Segmentation}
\authorrunning{M. Zhen, S. Li, L. Zhou, J. Shang, H. Feng, T. Fang and L. Quan}
%
\author{Mingmin Zhen\inst{1}\orcidID{0000-0002-8180-1023} \and  Shiwei Li\inst{2} \and Lei Zhou\inst{1}  \and \\ Jiaxiang Shang\inst{1} \and Haoan Feng\inst{1} \and Tian Fang\inst{2} \and Long Quan\inst{1}}
\institute{Hong Kong University of Science and Technology \\
	\email{\{mzhen,lzhouai,hfengac,jshang,quan\}@cse.ust.hk} \and
	Everest Innovation Technology \\
	\email{\{sli,fangtian\}@altizure.com}
}

\maketitle

\begin{abstract}
In this paper, we introduce a novel network, called discriminative feature network (DFNet), to address the unsupervised video object segmentation task. To capture the inherent correlation among video frames, we learn  discriminative features (D-features) from the input images that reveal feature distribution from a global perspective. The D-features are then used to establish correspondence with all features of test image under conditional random field (CRF) formulation, which is leveraged to enforce consistency between pixels.  The experiments verify that DFNet outperforms state-of-the-art methods by a large margin with a mean IoU score of 83.4\%  and ranks first on the DAVIS-2016 leaderboard while using much fewer parameters and achieving  much more efficient performance in the inference phase. We further evaluate DFNet on the FBMS dataset and the video saliency dataset ViSal, reaching a new state-of-the-art. To further demonstrate the generalizability of our framework, DFNet is also applied to the image object co-segmentation task. We perform experiments on a challenging dataset PASCAL-VOC  and observe the superiority of  DFNet. The thorough experiments verify that DFNet is able to capture and mine the underlying relations of images and discover the common foreground objects. 
\end{abstract}

\section{Introduction}

The research on video object segmentation (VOS),  which aims to  separate primary foreground objects from their background in a given video, is often divided into two categories, \emph{i.e.}, semi-supervised and unsupervised setting. The semi-supervised VOS (SVOS) provides a mask of the first frame, which can be taken as the prior knowledge about the target in subsequent frames.  By comparison, unsupervised VOS  (UVOS) is in general more challenging, as it requires a further step to distinguish the target object from a complex and diverse background without prior information. In this paper, we focus on the latter challenging issue.\\
\indent Recently, several works, such as COSNet \cite{vos_cosnet}, AGNN \cite{vos_agnn} and AnDiff \cite{vos_andiff},  model the long-term correlations  between frames to explore global information  inspired by the non-local operation introduced by Wang et al. \cite{nonlocal}. However, the limitations are obvious as the computation requirement is very high, especially for AGNN \cite{vos_agnn}. Besides,  the local consistency cues are overlooked, which is  essential for UVOS task. \\
\indent  Motivated by the above observations, we propose a discriminative feature learning network, which is denoted as DFNet, to model the long-term correlations between video frames. Specifically, DFNet takes several frames from the same video as input and learns the discriminative features, which can denote the whole feature distribution of the input frames. The feature map for each frame is correlated with these discriminative features under CRF formulation, which is used to boost the smoothness and consistency of similar pixels.  The proposed approach is advantageous to mine the discriminative representation from a global perspective, while at the same time helps to capture the rich contextual information within video frames. DFNet is sufficiently flexible to process variable numbers of input frames during inference, enabling it to consider more input information and gain better performance.\\
\indent  To verify the proposed method, we extensively evaluate DFNet on two widely-used video object segmentation datasets, namely DAVIS16 \cite{davis_16} and FBMS \cite{fbms}, showing its superior performance over current state-of-the-art methods. More specifically, DFNet ranks first  on the DAVIS-2016 leaderboard with a mean IoU score of 83.4\%, which is 1.7\% higher than state-of-the-art method \cite{vos_andiff}. DFNet also achieves state-of-the-art results on FBMS \cite{fbms} and the ViSal \cite{video_gafl} video saliency benchmark. To further demonstrate its advantages and generalizability, we apply DFNet to image object co-segmentation task, which aims to extract the common objects from a group of semantically related images. It also gains better results on the representative dataset PASCAL VOC  \cite{coseg_faktor13} over previous methods.\\

\section{Related work}
\indent \textbf{Unsupervised Video Object Segmentation}
 Recently, there are many works for UVOS task, which  focus on the fully convolutional neural network based  models.  MPNet \cite{vos_motion_pattern}, a purely optical flow-based method,  discards appearance modeling and casts segmentation as foreground motion prediction, which poorly deals with static foreground objects.  To better address this problem, several methods \cite{vos_vis_memory,segflow,motadapt,mbn} suggest adopting two-stream fully convolutional networks, which fuse the motion and appearance information for object inference. In \cite{vos_vis_memory},  a convolutional gated recurrent unit is employed to extend the horizon spanned by optical flow based features. Li et al. \cite{mbn} attempt to address this issue by employing a bilateral network for detecting the motion of background objects. RNN based methods are also a popular choice.  Song et al. \cite{vos_pdb} propose a novel convolutional long short-term memory \cite{lstm} architecture, in which two atrous convolution \cite{deeplabv3} layers are stacked along the forward axis and propagate features in opposite directions. COSNet \cite{vos_cosnet} adopts a gated co-attention mechanism to model the correlation of input video images.  In AGNN \cite{vos_agnn}, a fully connected graph is built to represent frames as nodes, and relations between arbitrary frame pairs as edges. The underlying pair-wise relations are described by a differentiable attention mechanism.   To exploit the correlations of images,  AnDiff \cite{vos_andiff} proposes a considerably simpler method, which propagates the features of the first frame (the “anchor”) to the current frame via an aggregation technique.\\
\noindent \textbf{Image Object Co-Segmentation} Different from UVOS, the image object co-segmentation task is to extract the common object with the same semantics from a group of semantic-related images.
 Recent researches \cite{coseg_hsu18,coseg_crf} use deep visual features to improve object co-segmentation, and they also try to learn more robust synergetic properties among images in a data-driven manner.   Hsu et al. \cite{coseg_hsu18} proposes a DNN-based method which uses the similarity between images in deep features and an additional object proposals algorithm \cite{coseg_proposal} to segment the common objects.  Yuan et al. \cite{coseg_crf} introduce a DNN-based dense conditional random field framework for object co-segmentation by cooperating co-occurrence maps, which are generated using selective search \cite{coseg_selectivesearch}. The very recent works \cite{coseg_semaware,coseg_deepcoseg}  propose end-to-end deep learning methods for co-segmentation by integrating the process of feature learning and co-segmentation inferring as an organic whole. By introducing the correlation layer \cite{coseg_deepcoseg} or a semantic attention learner \cite{coseg_semaware}, they can utilize the relationship between the image pair and then segment the co-object in a pairwise manner. In \cite{coseg_li19}, a recurrent network architecture is proposed to address group-wise object co-segmentation. 
\begin{figure*}[t!]
	\centering
	\includegraphics[width=0.98\linewidth]{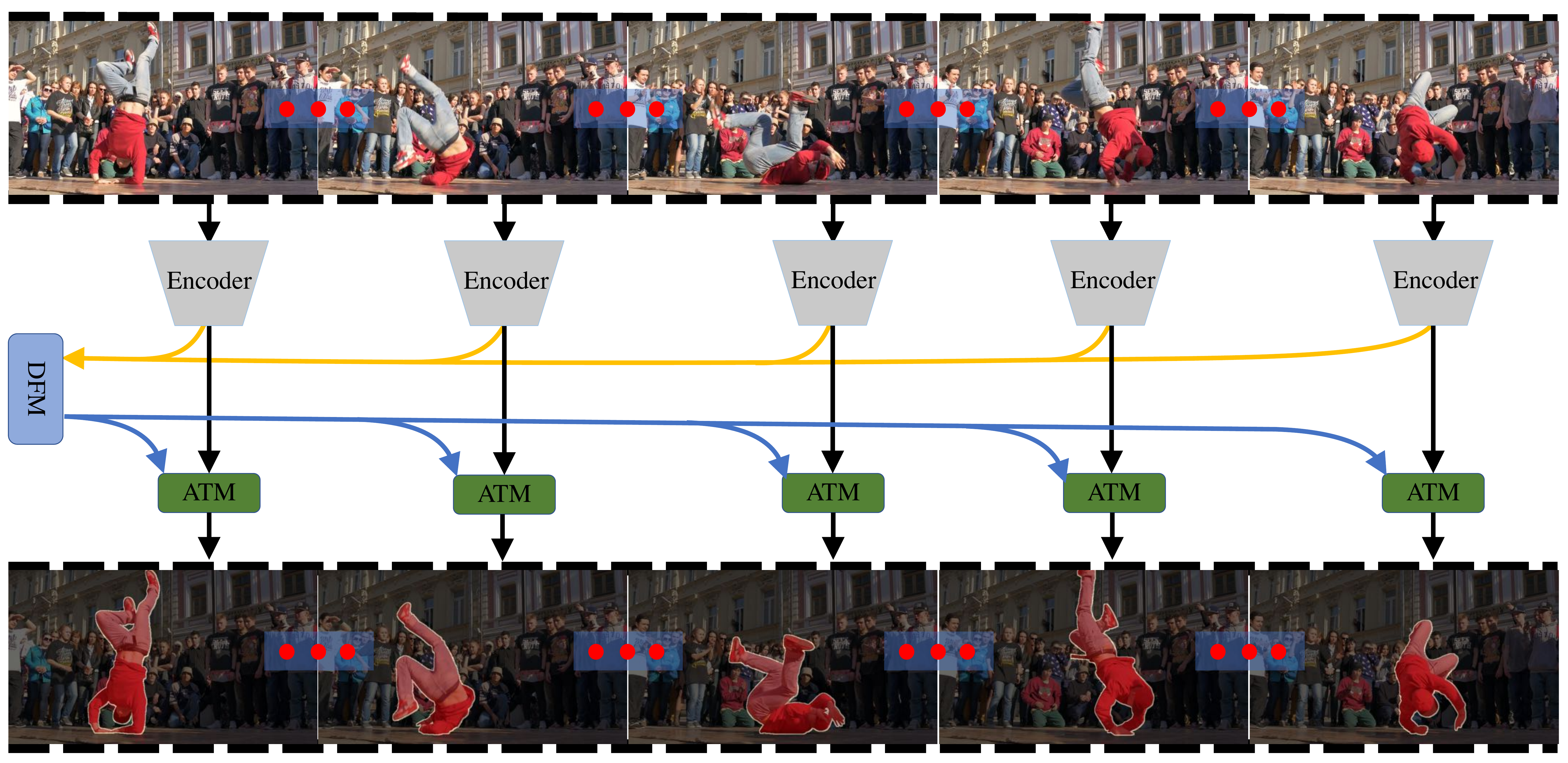}
	\caption{Overall pipeline of the proposed method. The features are first obtained from the encoder module and goes through the discriminative feature module (DFM) to extract  discriminative features. The discriminative features are then used by attention module (ATM) to recontruct a new feature map, which is used to correlate the input frames.}
	\label{vos_video}
\end{figure*}
\section{The Proposed Method}
\indent In this section, we present the proposed DFNet in detail, which is illustrated in Figure \ref{vos_video}. We first give an overview of the whole architecture in section \ref{sec_vos_architecture}. Next, the discriminative feature module (DFM), which captures the global feature distribution of all input images, is elaborated in section \ref{sec_dfm}. Then we introduce the attention module (ATM) in section \ref{sec_atm}, which reconstructs a new feature map modeling the long-term dependency. 
\subsection{Network Architecture}\label{sec_vos_architecture}
For the UVOS task, the target object in the given video images can be deformed and occluded, which often deteriorates the performance of estimated binary segmentation results. To recognize the target object, our method should be of two essential properties: (i) the ability to extract foreground objects from the individual frame; (ii) the ability to keep consistency among the video frames. To achieve these two goals, we correlate the features of each input image with discriminative features, which is extracted from  input images selected from the same video randomly. \\
\indent As shown in Figure \ref{vos_video}, we present the proposed network architecture in detail. The proposed network takes several images as input. The shared feature encoder, which adopts the fully convolutional DeepLabv3 \cite{deeplabv3}, extracted the features from the input images. The obtained feature maps are then fed into a $1 \times 1$ convolutional layer to reduce the feature map channel to 256, and the output feature maps for all input images are taken as input for the discriminative feature module (DFM), which extract the discriminative features (D-features).  The input feature for each image and the D-features go through an attention module (ATM) to reconstruct a new feature map and then one $3 \times 3$ convolutional layer followed by ReLU, batch normalization (BN) layer and one $1 \times 1$ convolutional layer followed by a $sigmoid$ operation are used to obtain final binary output. \\
\begin{figure*}[t!]
	\centering
	\includegraphics[width=0.75\linewidth]{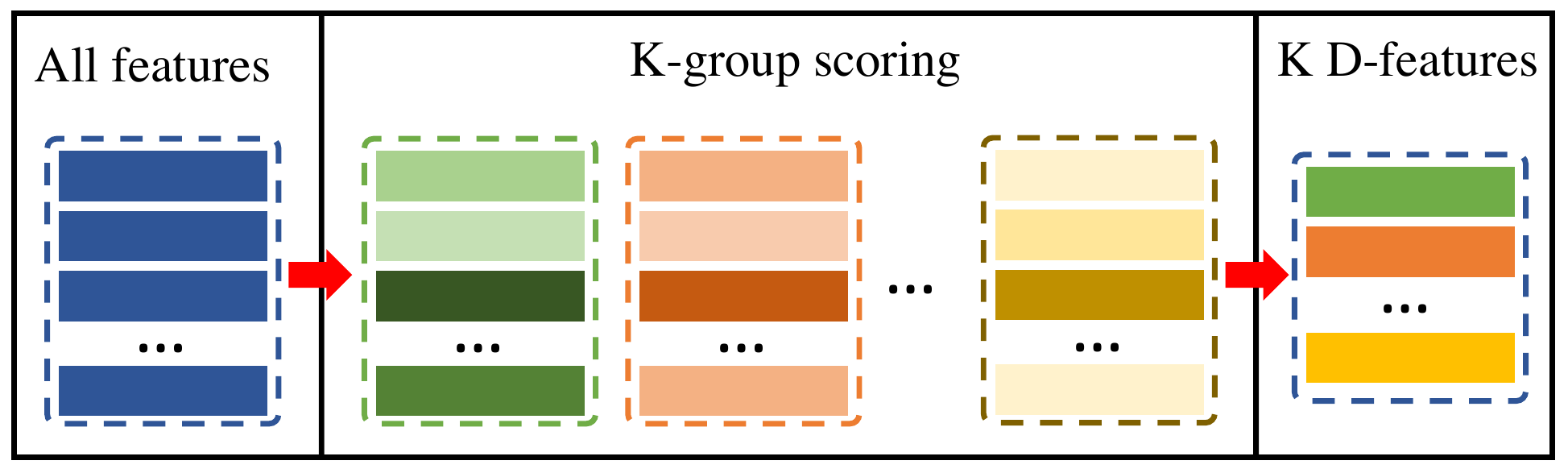}
	\caption{Illustration of DFM. The features from input images are first reshaped into one-dimensional vectors. The K-group scoring module is adopted to score the features. Based on the K-group scores, we can obtain final K-D features. The details are presented in Section \ref{sec_dfm}.}
	\label{vos_dfm}
\end{figure*}
\indent More formally,  given a set of input frames $\mathcal{I}=\{I_i \in \mathbb{R}^{H \times W \times 3}\}_{i=1}^{\mathcal{N}}$, we want to segment out the binary masks $\mathcal{S}=\{S_i \in \{0, 1\}^{H \times W}\}_{i=1}^{\mathcal{N}}$ for all frames. The features extracted from DeepLabv3 are denoted as $\mathcal{F}=\{F_i \in \mathbb{R}^{h \times w \times c}\}_{i}^{\mathcal{N}}$, where $h \times w$ indicates the spatial resolution of feature map and $c$ represents the feature map channels. Since we follow the original deepLabv3, which employs dilated convolution, the output feature map $F_i$ is  $\frac{1}{8}$   smaller than the input image $I_i$. 
\subsection{Discriminative Feature Module}\label{sec_dfm}
We learn the discriminative features from the features of all input images. Specifically, all feature maps $\mathcal{F}$ from the input images are first concatenated to form a large feature map with size $\mathcal{N} \times h \times w \times c$ and then reshaped as $F^{a} \in \mathbb{R}^{\mathcal{N}hw \times c}$. As shown in Figure \ref{vos_dfm}, we then use a K-group scoring module to obtain K-group scores, which is used to distinguish the discriminative features from the noisy features. For each scoring group,   a  weight matrix $\mathcal{W}_k \in \mathcal{R}^{c \times 1}$  and $F^{a}$ is multiplied to get a initial score result with size $\mathcal{N}hw \times 1$.  We apply a softmax function to calculate the final scores:
\begin{equation}
s_i^{k} = \frac{exp (F^{a}_{i} . W_k)}{\sum_{i}^{\mathcal{N}hw} exp(F^a_{i} . W_k)}
\end{equation}
where $s_i^{k}$ is the score for $i^{th}$ feature of $F^a$ and  measures the discriminability of the feature. The final discriminative feature for $k^{th}$ scoring group is computed as $	F_{k}^{d} = \sum s_i^{k}F^{a}_{i}$. By this way, we can obtain $K$ discriminative features $F^{d} \in \mathbb{R}^{K \times c}$.\\
\indent The $K$ D-features are used to describe the feature distribution from a global perspective. The key of the  D-features computation is the  scoring weight $\mathcal{W}_k$. In our training step, we initialize the $\mathcal{W}_k$ by using Kaiming’s initialization method \cite{kaiming_initialization}.  For each updating iteration, we adopt the  moving averaging mechanism, which is used in  batch normalization (BN) \cite{batch_normalization}. After obtaining the D-feature $F_k^{d}(t)$ at training step $t$, we update the $\mathcal{W}_k$ as:
\begin{equation}
\mathcal{W}_k(t) = \lambda \mathcal{W}_k(t - 1) + (1 - \lambda) F_k^{d}(t)\label{mving_average}
\end{equation}
where $\lambda$ is the momentum. In our experiments, we set it to 0.5. As we train our network on a multiple-GPU machine, we also adopt the synchronized weight updating strategy motivated by synchronized BN \cite{inplaceabn}. Specifically, the images from the same video sequence are fed into the network on one GPU. Thus, we will get different D-features $F_k^{d}(t)$ at step $t$ for different GPUs. For the synchronized processing, we sum up these D-features $F_k^{d}(t)$ across GPUs and compute the average feature $\overline{F}_k^{d}(t)$, which will be used in Equation \ref{mving_average}.  The updated $\mathcal{W}_k$ is synchronized on all GPUs. The whole computation is differentiable and trainable. In the inference step, the weight $\mathcal{W}_k$ is kept fixed, which is similar to BN operation.
\subsection{Attention Module with CRF}\label{sec_atm}
\begin{figure*}[t!]
	\begin{minipage}[t]{0.65\linewidth} 
		\centering
		\includegraphics[width=0.99\linewidth]{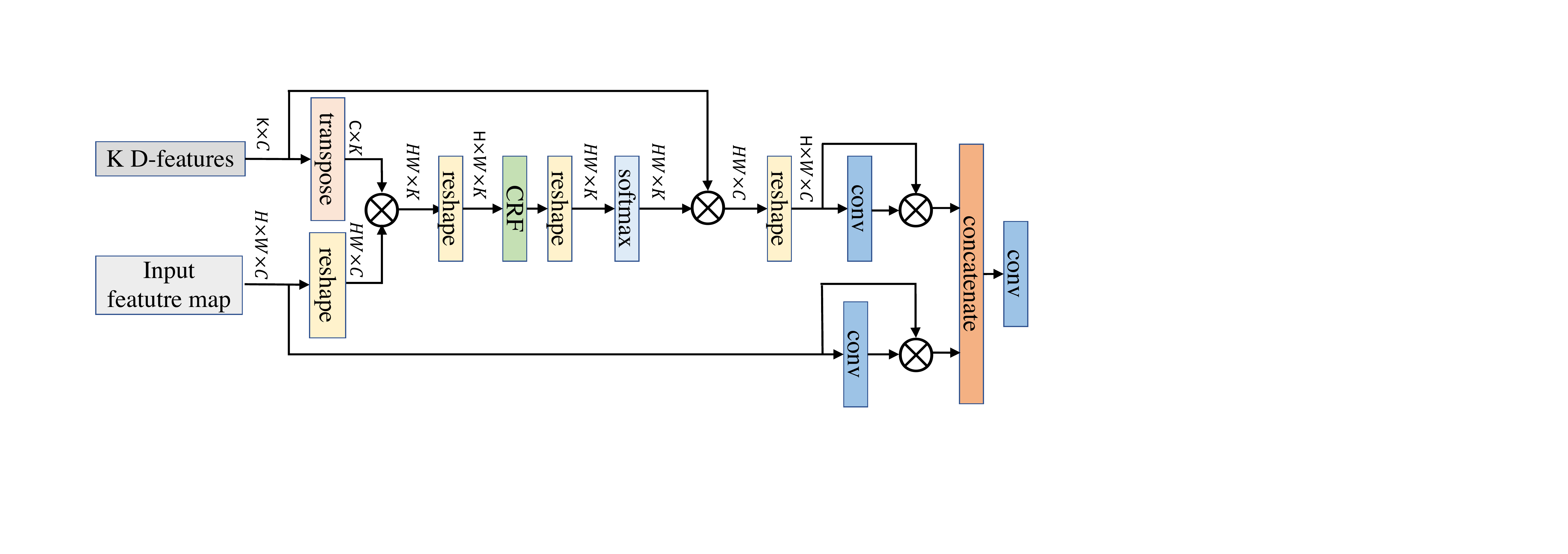}
		\centerline{(a)}
	\end{minipage}
	\begin{minipage}[t]{0.34\linewidth} 
		\centering
		\includegraphics[width=0.99\linewidth]{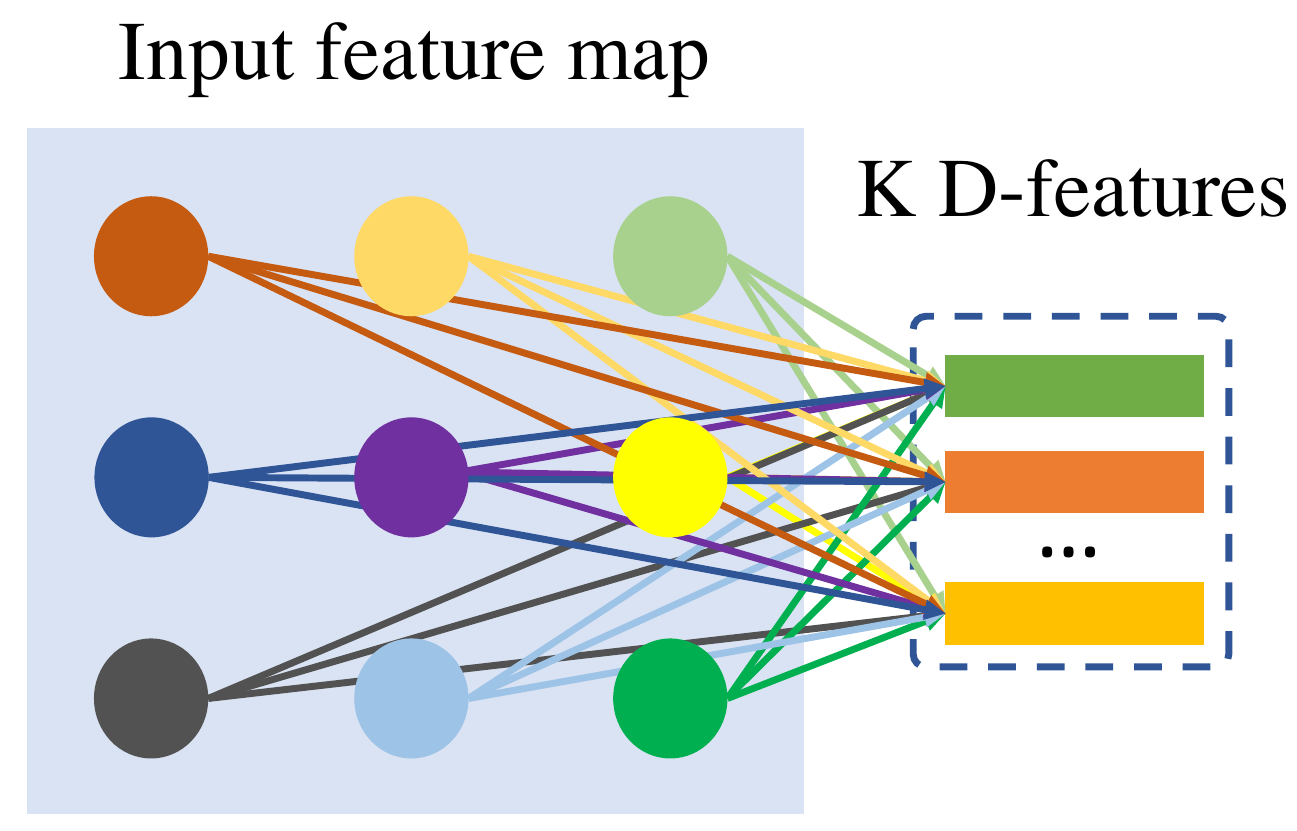}
		\centerline{(b)}
	\end{minipage} 
	\caption{(a) Illustration of ATM. The input feature map and $K$ D-features are correlated to model long-term dependency; (b)  Illustration of attention mechanism. The details are presented in Section \ref{sec_atm}.}
	\label{vos_atm}
\end{figure*}
To model the long-term dependency, we adopt the attention module to correlate input image and the discriminative features.  For the obtained $K$ D-features $F^{d} \in \mathbb{R}^{K \times c}$ and the feature map $F_i \in \mathbb{R}^{h \times w \times c}$ of input image, we follow \cite{vos_cosnet,vos_agnn} to compute the attention matrix $P \in \mathbb{R}^{hw \times K}$ as shown in Figure \ref{vos_atm} (a). Specifically, we obtain $P$ from $F^{d}$ and $F_i$ as follows:
\begin{equation}
P = reshape(F_i)W_{att}transpose(F^{d})
\end{equation}
where $W_{att} \in \mathbb{R}^{c \times c}$ is a learnable weight matrix. The  D-feature matrix $F^d$ are tranposed with size $c \times K$ and feature map $F_i$ is reshaped with size $hw \times c$. For the obtained attetnion matrix $P$, each element indicates the similarity of the corresponding feature of $F_i$ and feature of $F^d$. As shown in Figure \ref{vos_atm} (b), the lines with different colors represent the similarity between input features and $K$ D-features. In previous attention methods \cite{vos_cosnet,vos_agnn,vos_andiff}, a new feature map is reconstructed based on the attention matrix by assigning $K$ D-features to input feature map as follows:
\begin{equation}
F^{new} = reshape(softmax(P)F^{d}) \label{new_feature}
\end{equation}
where the new feature map $F^{new}$ is of size $h \times w \times c$. \\
\indent The attention map computation can also be considered as multi-label classification problem and the assignment of  D-features corresponds to a different label. Our intuition is that neighboring pixels in the same local region tend to have similar labels ($K$ D-features), and pixels near borders or edges may have significantly different labels. We regard the reshaped attention map with size $h \times w \times K$ as fully connected pairwise conditional random fields conditioned on the corresponding image $I$, in which each pixel is to be assigned with a D-feature for reconstructing the new feature map.\\
\indent Let $\textbf{x} = \{x_1, x_2, ..., x_M\}$ be the label vector of $M$ pixels in the reshaped attention map. Component $x_i$ belongs to $\{1, 2, ..., K\}$ where $K$ is the number of labels (D-features). The probability of the label assignment is defined in the form of Gibbs distribution as $P(\textbf{x}|\textbf{I}) = \frac{1}{Z}exp(-E(\textbf{x}|\textbf{I}$)), where $E(x)$ is the energy function which describes the cost of label assigning and $Z$ is a normalization factor. For convenience we drop the notation of condition $\textbf{I}$ in the followings. Following the formulation of  \cite{densecrf}, the energy function is defined as 
\begin{equation}
E(\textbf{x}) = \sum_{i=1}^{M}\psi_{u}(x_i) + \sum_{i < j}\psi_{p}(x_i, x_j)
\end{equation}
where the unary energy components $\psi_{u}(x_i)$ measure the cost of the pixel $i$ taking the label $x_i$, and pairwise energy components $\psi(x_i, x_j)$  measure the cost of assigning labels $x_i, x_j$ to pixels $i, j$ simultaneously. In our formulation, unary energies are set to be reshaped attention map $P$, which predicts labels for pixels without considering the smoothness and the consistency of the label assignments. The pairwise energies provide an image data-dependent smoothing term that encourages assigning similar labels to pixels with similar properties. \\
\indent The CRF model can be implemented in neural networks as shown in \cite{CRFasRNN,crf_conv}, thus it can be naturally   integrated in our network, and optimized in the end-to-end training process. After the CRF module, we can obtain a refined attention map which takes the smoothness and consistency  into consideration. We follow Equation \ref{new_feature} to reconstruct a new feature map. \\
\indent We also adopt a self-weight method to weight the new feature map $F^{new}$ and input feature map $F_i$. The self-weight is formulated as follows:
\begin{equation}
F^{new} = F^{new}* conv(F^{new}), F_i = F_i* conv(F_i)
\end{equation}
where we use $1 \times 1$ convolutional layer to get the weight, which indicates the importance of features in the feature map. At last, we concatenate the feature map $F^{new}$ and $F_i$ and feed the obtained feature map into the convolutional layers to get binary  segmentation results.

\section{Experiments}
We first report performance on the unsupervised video object segmentation task in Section \ref{exp_uvos}. Then, in Section \ref{exp_iocs}, to further demonstrate the advantages of the proposed model, we test it on  image object co-segmentation task. At last, we conduct an ablation study in Section \ref{exp_ablation} and model analysis in Section \ref{exp_analysis}
\begin{table*}[t!]
	\caption{Quantitative results on the test set of DAVIS16, using the region similarity $\mathcal{J}$, boundary accuracy $\mathcal{F}$ and time	stability $\mathcal{T}$.  For FBMS dataset, we report the F-measusre results. The best scores are marked in bold.}
	\centering
	\setlength{\tabcolsep}{1.4mm}{
		\begin{tabular}{l|l|ccc|c}
			\toprule
			&&\multicolumn{3}{|c|}{DAVIS}&FBMS\\
			Method &\multicolumn{1}{|c|}{Year} &  $\mathcal{J}$   Mean$\uparrow$  & $\mathcal{F}$ Mean$\uparrow$& 	$\mathcal{T}$ Mean$\downarrow$  &F-measure\\
			\hline
			TRC  \cite{vos_trc} &CVPR12& 47.3  & 44.1  &39.1 &-\\
			CVOS \cite{vos_cvos}&CVPR15 &48.2& 44.7&25.0&- \\
			KEY \cite{vos_key} &ICCV11 & 49.8  &42.7  &26.9 &-\\
			MSG \cite{vos_msg} &ICCV11 & 53.3  & 50.8&30.2 &-\\
			NLC \cite{vos_nlc} & BMVC14 & 55.1  &52.3 &42.5&- \\
			CUT \cite{vos_cut} & ICCV15 & 55.2 &55.2 & 27.7 &-\\
			FST \cite{vos_fst}&ICCV13 & 55.8 &51.1  &36.6&69.2\\
			ELM \cite{vos_elm} &ECCV18 &61.8 &61.2	 &25.1&-\\
			TIS \cite{vos_tis}&WACV19&62.6 &59.6&33.6&-\\
			SFL \cite{segflow} & ICCV17 & 67.4 &66.7  &28.2&- \\
			LMP \cite{vos_motion_pattern} &CVPR17 & 70.0 &65.9  &57.2&77.5\\
			FSEG \cite{vos_fusionseg} &CVPR17 & 70.7  &65.3  &32.8 &-\\
			UOVOS \cite{vos_uovos}  &TIP19        &73.9        &68.0       &39.0          &-\\ 
			LVO \cite{vos_vis_memory} &ICCV17 & 75.9  &72.1& 26.5&77.8 \\
			ARP \cite{vos_region_aug} &CVPR17 & 76.2  &70.6  &39.3 &-\\
			PDB \cite{vos_pdb} & ECCV18&  77.2 &74.5 &29.1 &81.5\\
			LSMO \cite{vos_lsmo} &IJCV19&78.2&	75.9&  21.2 &-\\
			MotAdapt \cite{motadapt} &ICRA19& 77.2&	77.4&   27.9&79.0\\
			EpO+ \cite{vos_epo}&WACV20&80.6	&	75.5& 19.3 &-\\
			AGS \cite{vos_ags}&		CVPR19 &	79.7&	77.4&	26.7   &-\\
			COSNet \cite{vos_cosnet}	&CVPR19		&80.5&	79.5	&	18.4 &-\\
			AGNN \cite{vos_agnn} & ICCV19 & 80.7 & 79.1&  33.7 &-\\
			AnDiff \cite{vos_andiff}&ICCV19		&	81.7	&80.5&	21.4&81.2\\
			\hline
			Ours &ECCV20 & \textbf{83.4} &\textbf{81.8} &\textbf{15.9}&\textbf{82.3}\\	
			\bottomrule
		\end{tabular}
	}	
	\label{davis16_sota}
\end{table*}
\begin{table*}[t!]
	\caption{Quantitative comparison results against  saliency methods using MAE and maximum F-measure on DAVIS16 \cite{davis_16}, FBMS \cite{fbms} and ViSal \cite{video_gafl}. The best scores are marked in bold. * means non-deep learning model.}
	\centering
	\setlength{\tabcolsep}{1.5mm}{
		
		\begin{tabular}{c|l|l|cc|cc|cc}
			\toprule
			
			& & & \multicolumn{2}{c|}{DAVIS16} &  \multicolumn{2}{c|}{FBMS} &  \multicolumn{2}{c}{ViSal} \\
			&Methods &\multicolumn{1}{|c|}{Year}& MAE$\downarrow$ & F$\uparrow$ & MAE$\downarrow$ & F$\uparrow$ & MAE$\downarrow$ & F$\uparrow$ \\
			\hline \hline 
			\multirow{11}{*}{Image }&Amulet \cite{sod_amulet} &ICCV17& 0.082 &69.9 & 0.110 &72.5 & 0.032 & 89.4 \\
			&SRM   \cite{sod_srm}  &ICCV17& 0.039 & 77.9 & 0.071 & 77.6 & 0.028 &89.0 \\
			&UCF   \cite{sod_ucf}   &ICCV17 &0.107 &71.6 &0.147 &67.9 &0.068 & 87.0 \\
			&DSS  \cite{sod_dss}    &CVPR17 &0.062 &71.7 & 0.083 & 76.4 & 0.028 &90.6 \\
			&MSR  \cite{sod_msr}   &CVPR17 & 0.057 & 74.6 & 0.064 & 78.7 &0.031 & 90.1 \\
			&NLDF  \cite{sod_nldf} &CVPR17 &0.056 & 72.3 & 0.092 &73.6 & 0.023 & 91.6 \\
			&DCL    \cite{sod_dcl} &CVPR16 &0.070 & 63.1 & 0.089 & 72.6 & 0.035 & 86.9 \\
			&DHS   \cite{sod_dhsnet}  &CVPR16 &0.039 &75.8 &0.083 &74.3 &0.025 &91.1 \\
			&ELD    \cite{sod_eld} &CVPR16    &0.070     &68.8    &0.103  &71.9  &0.038 &89.0 \\
			&KSR   \cite{sod_kernelized}  &ECCV16 &0.077  &60.1 &0.101   &64.9   &0.063 &82.6 \\
			&RFCN \cite{sod_rfcn}&ECCV16  &0.065 &71.0 &0.105  &73.6  &0.043 &88.8  \\
			\hline
			\multirow{8}{*}{Video }&FGRNE \cite{video_fgrne}&CVPR18 & 0.043 &   78.6 & 0.083 & 77.9 & 0.040 & 85.0 \\
			&FCNS   \cite{video_fcns}& TIP18   & 0.053 & 72.9 & 0.100 & 73.5 & 0.041 &87.7 \\
			&SGSP* \cite{video_sgsp}&TCSVT17 & 0.128 & 67.7 & 0.171 & 57.1 &0.172 &64.8 \\
			&GAFL* \cite{video_gafl}  &TIP15  &0.091 &57.8 &0.150 &55.1 & 0.099 &72.6 \\
			&SAGE* \cite{video_sage} &CVPR15   &0.105  &47.9 &0142 &58.1  &0.096 &73.4  \\
			&STUW* \cite{video_stuw} &TIP14 &0.098& 69.2 &0.143 &52.8 &0.132 &67.1\\
			&SP*  \cite{video_sp}  &TCSVT14  &0.130  & 60.1  &0.161  &53.8  &0.126  &73.1  \\
			&PDB  \cite{vos_pdb} &ECCV18    &0.030 &84.9   &0.069 &81.5 &0.022 &91.7 \\
			&AnDiff \cite{vos_andiff}&ICCV19  & 0.044 &80.8 &0.064 & 81.2 & 0.030 & 90.4 \\
			\hline
			& Ours &ECCV20  & \textbf{0.018} & \textbf{89.9} & \textbf{0.054} &\textbf{ 83.3} & \textbf{0.017} &\textbf{92.7}\\ 				
			
			\bottomrule
		\end{tabular}
	}	
	
	\label{visal_sota}
\end{table*}
\subsection{Unsuperviesed Video Object Segmentation Task}\label{exp_uvos}
\textbf{Dataset and Evaluation Metric} To evaluate UVOS task, a golden dataset \textbf{DAVIS16} is often used  \cite{vos_cosnet,vos_agnn,vos_ags,vos_lsmo,vos_pdb}. DAVIS16 is a recent dataset which consists of 50 videos in total (30 videos for training and 20 for testing). Per-frame pixel-wise annotations are offered. For quantitative evaluation, following the standard evaluation protocol from \cite{davis_16}, we adopt three metrics, namely region similarity $\mathcal{J}$, which is the intersection-over-union of the prediction and ground truth, boundary accuracy $\mathcal{F}$, which is the F-measure defined on contour points in the prediction and ground truth, and time stability $\mathcal{T}$, which measures the smoothness of evolution of objects across video sequences. \textbf{FBMS} \cite{fbms} is comprised of 59 video sequences. Different from the DAVIS16 dataset, the ground-truth of FBMS is sparsely labeled (only 720 frames are annotated). Following the common setting \cite{vos_pdb,vos_andiff,motadapt}, we validate the proposed method on the testing split, which consists of 30 sequences. On the FBMS dataset, the F-measure is used as evaluation metric.  We also follow  \cite{vos_pdb,vos_andiff} to report saliency evaluations of our method on DAVIS, FBMS and a video salient object detection dataset ViSal \cite{video_gafl} for demonstrating the robustness and wide applicability of our method.  The \textbf{ViSal} \cite{video_gafl} dataset is a video salient object detection benchmark. The length of videos in ViSal ranges from 30 to 100 frames, and totally 193 frames are manually annotated. The whole ViSal dataset is used for evaluation. We report the mean absolute error (MAE) and the F-measure on the three datasets. \\
\noindent \textbf{Implementation Details} In the \textbf{training} step, following \cite{vos_cosnet,vos_agnn,vos_pdb},  we use both static data from image salient object segmentation datasets, MSRA10K \cite{msra10k}, DUT \cite{duts}, and video data from the training set of DAVIS16 to train our model. The training process is divided into two steps. First, we use the static training data to train our backbone encoder (DeepLabV3) to extract more discriminative foreground features. The learning rate is set to 0.01 and the batch size is 12.  Then we use the DAVIS16 training data to train the whole model with learning rate of 0.001. The batch size is set to 8. For each video sequence, we follow \cite{vos_andiff} to select the first frame as an anchor and randomly sample one image as the training example. The model is trained with binary cross-entropy loss. Network parameters are optimized via stochastic gradient descent with weight decay 0.0001. We adopt the ``poly'' learning rate policy where the initial learning rate is multiplied by $(1-\frac{iter}{max\_iter})^{power}$ with $power=0.9$. Raw predictions are upsampled via bilinear interpolation to the size of the ground-truth masks. In the \textbf{inference} step, multiscale and mirrored inputs are employed to enhance the final performance.	The final heatmap is the mean of all output heatmaps. Thresholding at 0.5 produces the final binary labels. We also follow \cite{vos_andiff} to adopt instance pruning as a post-processing method.\\
\begin{figure*}[t!]
	\begin{minipage}[t]{0.32\linewidth} 
		\centering
		\includegraphics[width=1\linewidth]{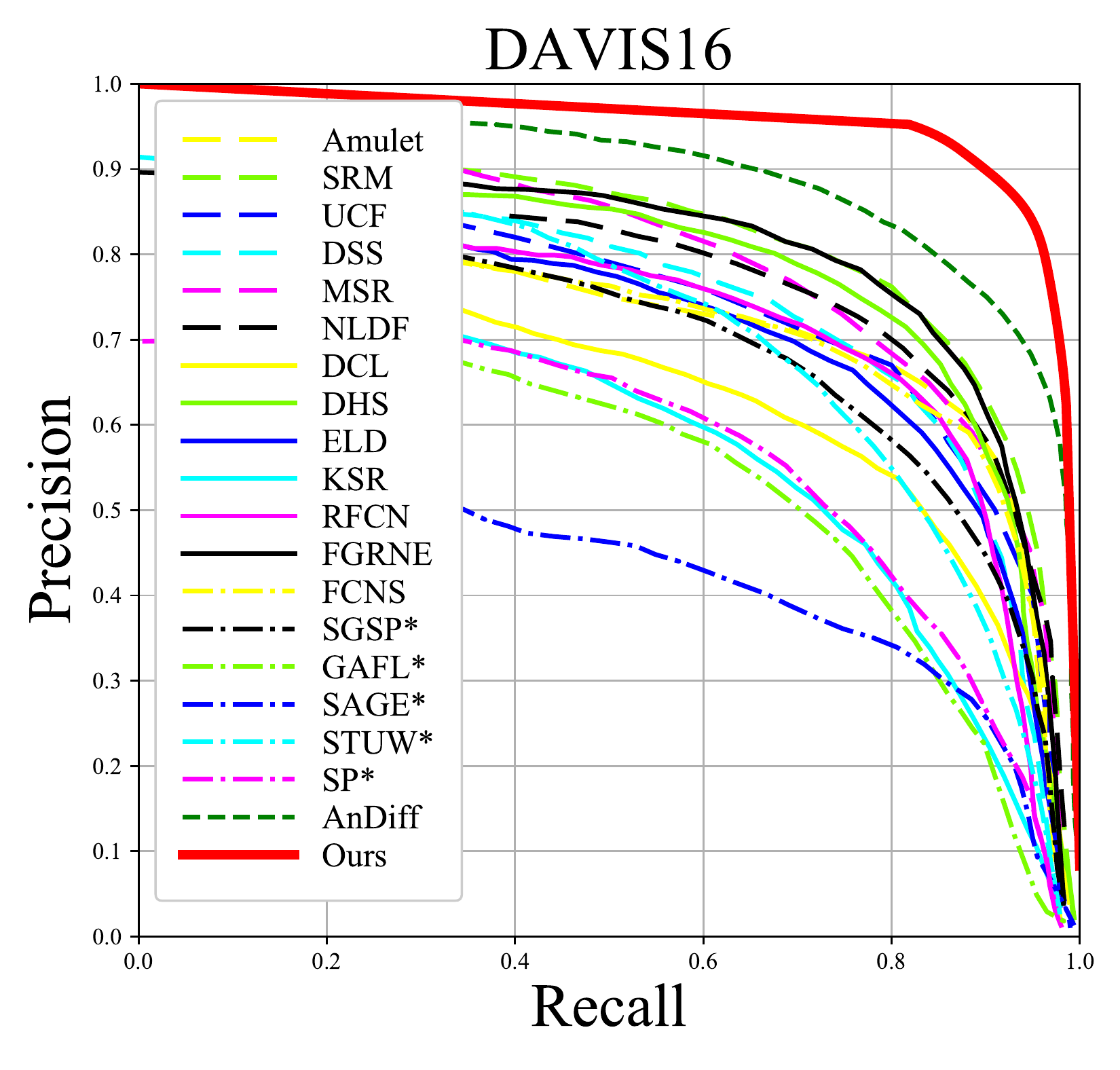}
	\end{minipage}
	\begin{minipage}[t]{0.32\linewidth} 
		\centering
		\includegraphics[width=1\linewidth]{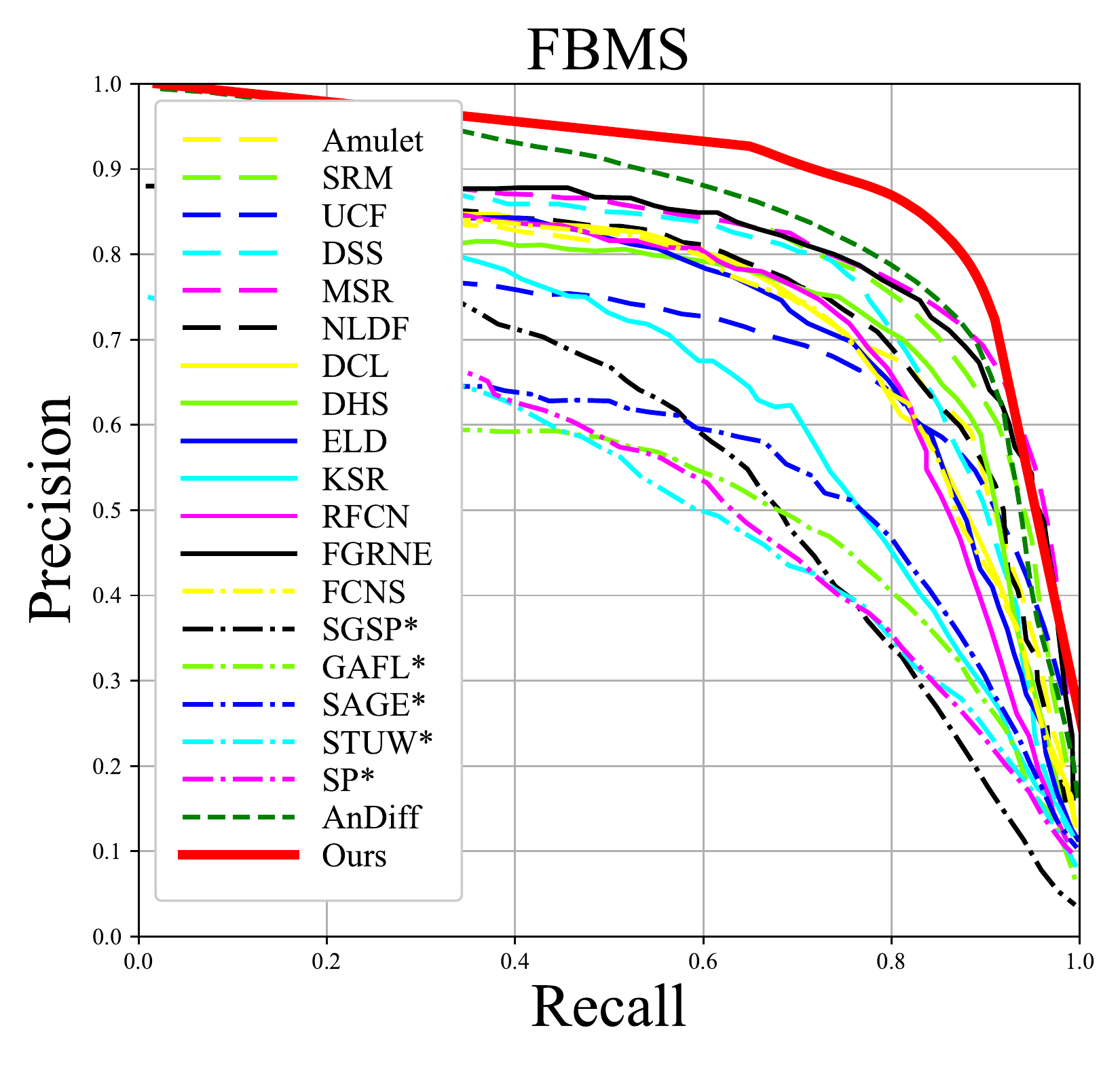}
	\end{minipage} 
	\begin{minipage}[t]{0.32\linewidth} 
		\centering
		\includegraphics[width=1\linewidth]{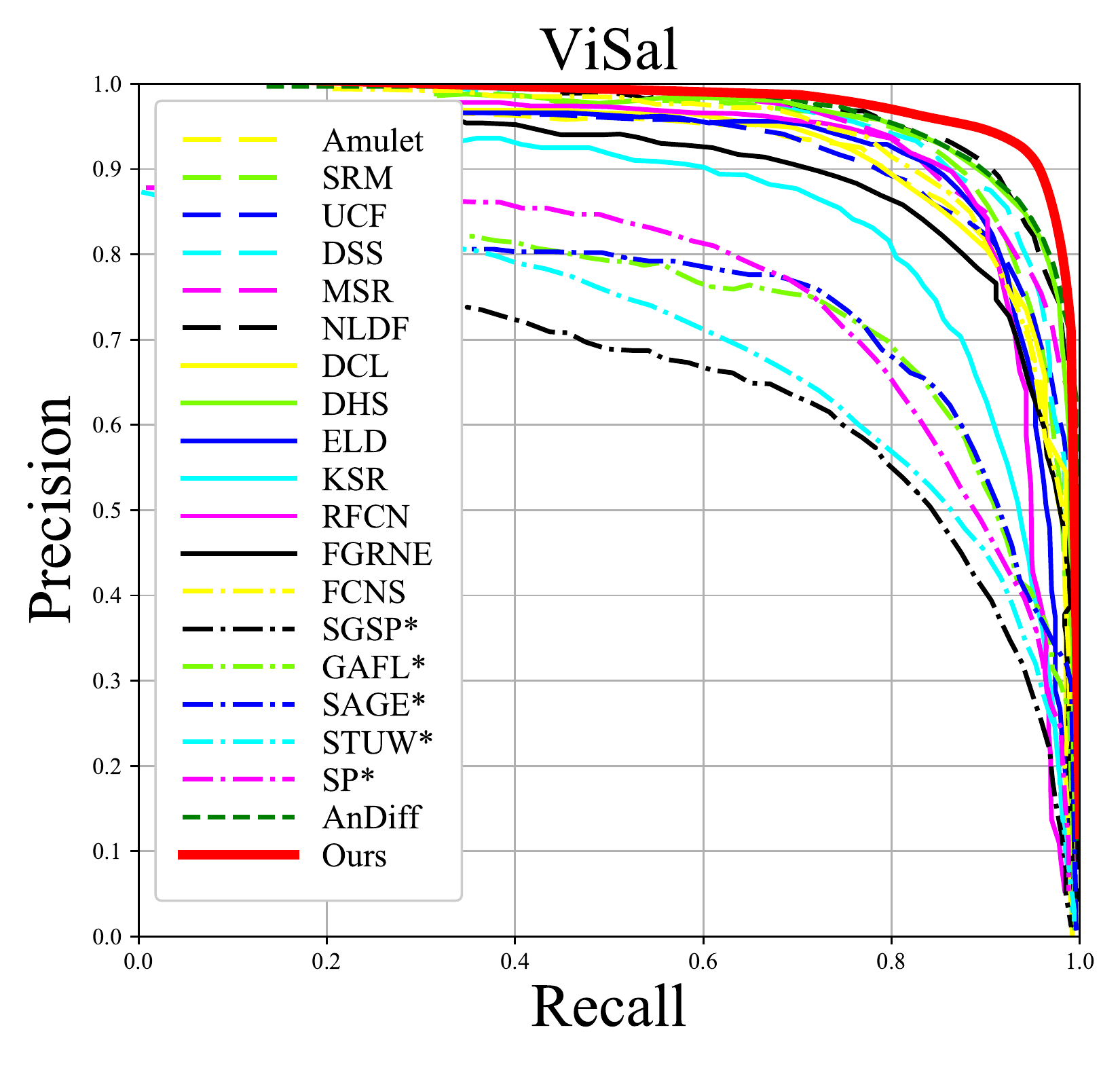}
	\end{minipage} 
	\caption{Quantitative comparison against other methods using PR curve on DAVIS16 \cite{davis_16}, FBMS \cite{fbms} and ViSal \cite{video_gafl} datasets. }
	\label{vos_pr}
\end{figure*}
\begin{figure*}[t!]
	\centering
	\includegraphics[width=0.95\linewidth]{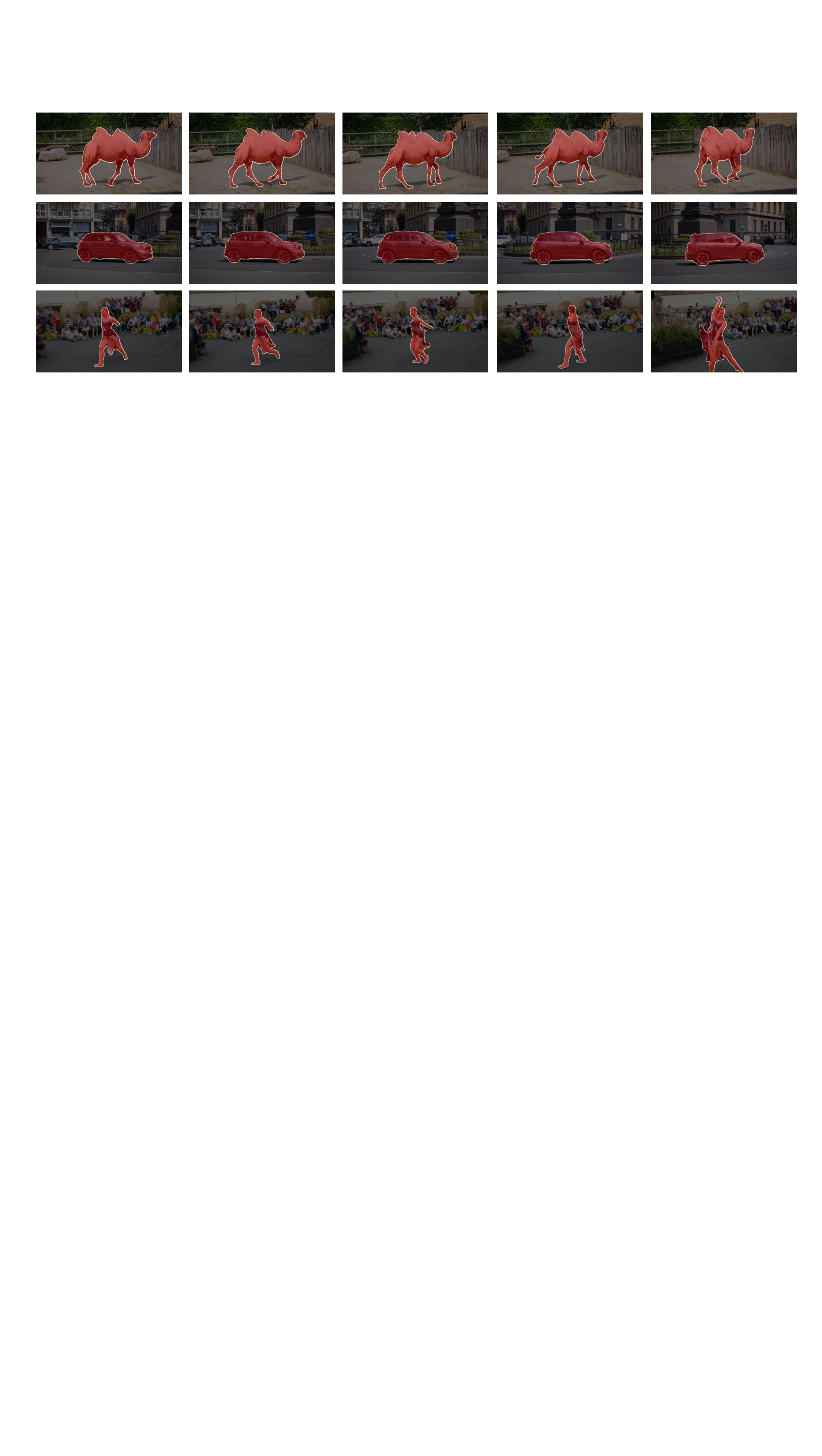}
	\caption{The visual results generated by our approach on the DAVIS16 dataset. From the first row to the last row, the corresponding
		video names are \emph{camel}, \emph{car-roundabout} and \emph{dance-twirl} respectively.}
	\label{davis_sample}
\end{figure*}
\begin{figure*}[t!]
	\centering
	\includegraphics[width=0.9\linewidth]{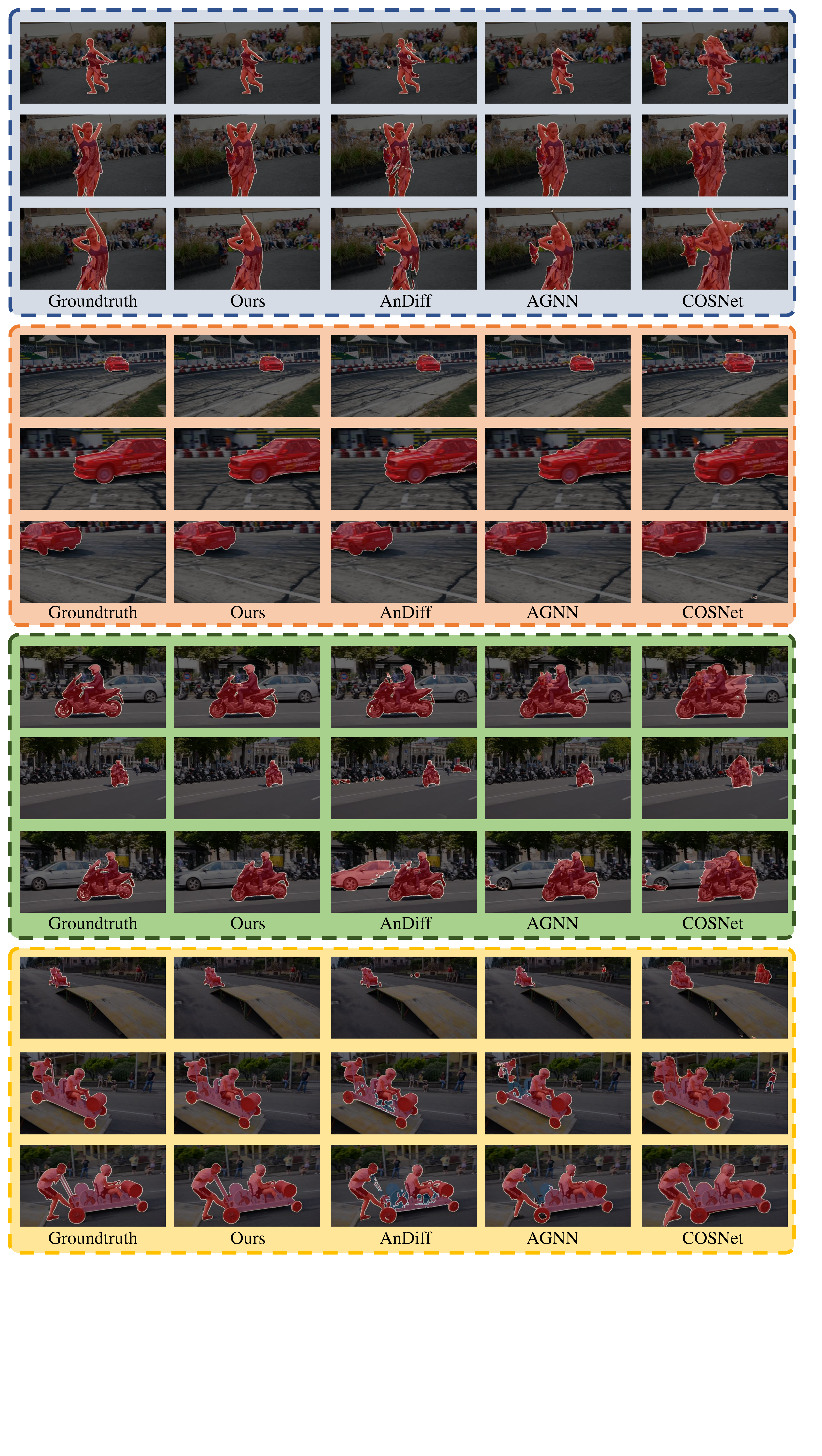}
	\caption{Qualitative comparison with state-of-the-art methods (AnDiff \cite{vos_andiff}, AGNN \cite{vos_agnn} and COSNet \cite{vos_cosnet}) on DAVIS16  dataset.   }
	\label{davis_cmp}
\end{figure*}
\noindent \textbf{Experimental Results } In Table \ref{davis16_sota}, we evaluate DFNet against state-of-the-art unsupervised VOS methods on the DAVIS16 public leaderboard. DFNet attains the highest performance among all unsupervised methods on the DAVIS16 validation set, while also achieving a new state-of-the-art on the FBMS test set. In particular, on DAVIS16 we outperform the second-best method (AnDiff \cite{vos_andiff}) by an absolute margin of 21.7\% in the  region similarity $\mathcal{J}$ and 1.3\% in the boundary accuracy $\mathcal{F}$. For the temporal stability $\mathcal{T}$, our method shows a more stable result over the video sequences by a large margin of 2.5  than the second-best method COSNet \cite{vos_cosnet}. We also outperform state-of-the-art method AnDiff \cite{vos_andiff} by 1.1\% in F-measure on the FBMS dataset.\\
\indent We also report the results on salient object detection for DAVIS16, FBMS and ViSal datasets as shown in Table \ref{visal_sota}. It can be observed that the proposed method improves state-of-the-art for all the three datasets for standard saliency scores, showing consistency with Table \ref{davis16_sota}. The largest improvements lie in DAVIS16, where both MAE and F-measure significantly outperform previous records. Especially for the F-measure, we outperform the second-best result by a significant margin of 9.1\%. The precision-recall analysis of DFNet is presented in Figure \ref{vos_pr}, where we demonstrate that our approach generally outperforms also existing salient object detection methods.
DFNet achieves superior performance in all regions of the PR curve on the DAVIS validation set, maintaining significantly higher precision at all recall thresholds. On the challenging FBMS test set, DFNet shows  inferior precision results than SP \cite{video_sp}at the recall threshold from 0.93 to 0.97 and FGRNE \cite{video_fgrne} from 0.94 to 0.95. But overall speaking,  DFNet maintains a clear advantage compared with all other methods. 
On the ViSal dataset, it is noteworthy that the precision is higher than the other methods at nearly all recall thresholds, except for the AnDiff \cite{vos_andiff} at the threshold from 0.98 to 0.99. All in all, the superiority of the proposed method is verified through the comparison of the PR curves.\\
\begin{table*}[t!]
	\caption{The performance of object co-segmentation on the PASCAL-VOC dataset under Jaccard index and Precision.  The numbers in red and green respectively indicate the best and the second best results. }
	\label{vos_voc}
	\resizebox{\textwidth}{!}{ %
		\centering
		\begin{tabular}{l|ccccccccccc|c}
			\toprule
			Method & Faktor13 \cite{coseg_faktor13} &Lee15 \cite{coseg_lee15}& Chang15 \cite{coseg_chang15}&Hati16 \cite{coseg_hati16} &Quan16 \cite{coseg_quan16}&Jerri.16 \cite{coseg_Jerripothula16}  &Wang17 \cite{coseg_wang17} & Jerri.17 \cite{coseg_Jerripothula17} &Han18 \cite{coseg_han18}&Hsu18 \cite{coseg_hsu18}&Li19 \cite{coseg_li19} &Ours
			\\ \hline\hline
			Avg. $\mathcal{P}$ &84.0 &69.8&82.4&72.5 & 89.0& 85.2 &84.3 &80.1&90.1 & 91.0  &\textcolor{green}{94.1} &\textcolor{red}{ 95.2} \\
			Avg. $\mathcal{J}$ &0.46&0.33 &0.29&0.25&0.52 &0.45& 0.52&0.40& 0.53 &0.60&\textcolor{green}{0.63} &\textcolor{red}{0.69}\\
			\bottomrule
		\end{tabular}
	}
\end{table*}
\begin{figure*}[t!]
	\centering
	\includegraphics[width=0.94\linewidth]{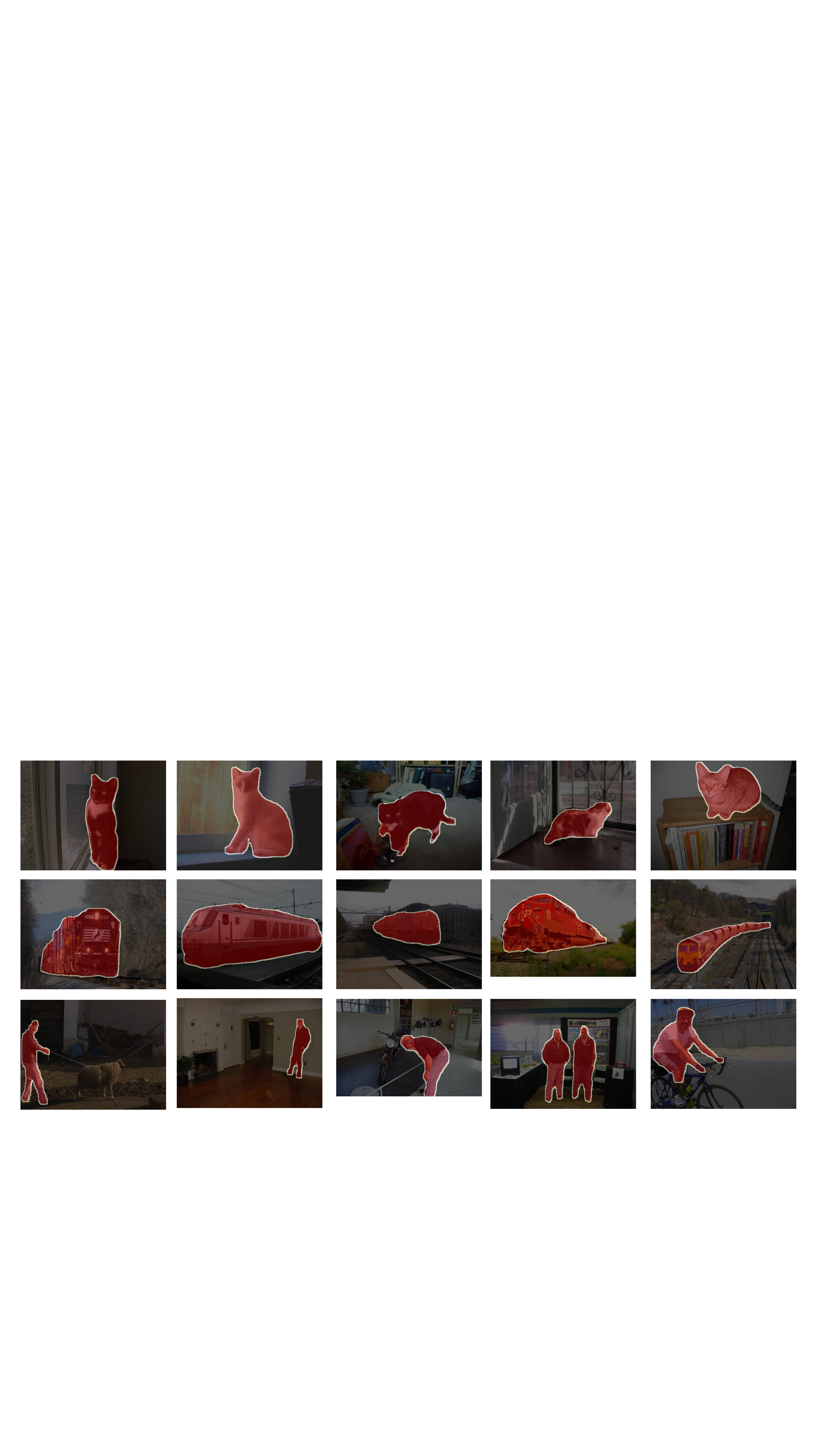}
	\caption{The co-segment results generated by our approach on the PASCAL-VOC dataset. From the first row to the last row, the
		classes are \emph{cat}, \emph{train} and \emph{person} respectively.}
	\label{pascal_sample}
\end{figure*}
\indent As shown in Figure \ref{davis_sample}, we visualize some qualitative results of the DAVIS16 dataset. We can see that the proposed
method can locate the primary region or target tightly by leveraging DFM and ATM with CRF to model long-term denpendency. The primary objects from the cluttered background are segmented out correctly. We also present the visual comparison results between DFNet and COSNet \cite{vos_cosnet}, AGNN \cite{vos_agnn} and AnDiff \cite{vos_andiff} in Figure \ref{davis_cmp}. In can be observed that the results of DFNet are more accurate and complete than the other three methods.
\subsection{Image Object Co-segmentation Task}\label{exp_iocs}
\textbf{Dataset and Evaluation Metric}  The PASCAL-VOC \cite{coseg_faktor13}  is a well-known dataset often used in image object co-segmentation task, which contains total 1,037 images of 20 object classes from PASCAL-VOC 2010 dataset. The PASCAL-VOC dataset is challenging and difficult due to extremely large intra-class variations and subtle figure-ground discrimination.  Following previous works \cite{coseg_faktor13,coseg_lee15,coseg_chang15,coseg_Jerripothula16,coseg_li19}, two widely used measures, precision ($\mathcal{P}$) and Jaccard index ($\mathcal{J}$), are adapted to evaluate the performance of object co-segmentation. \\
\noindent \textbf{Implementation Details} We follow \cite{coseg_groupsem,coseg_li19} to train the proposed network  with generated training data from the existing MS COCO dataset \cite{mscoco}. The learning rate is set to 0.01 and batch size is 12. For each group images, we randomly select three images as one training example. Other training setups are the same as those in previous unsupervised VOS task. After training, we evaluate the performance of our method on the PASCAL VOC dataset. When processing an image, we leverage another 4 images belonging to the same group to form a subgroup as inputs. We adopt a threshold 0.5 to generate final binary  masks.\\
\noindent \textbf{Experimental Results}  We compare our methods with state-of-the-art methods on the PASCAL VOC dataset. As shown in Table \ref{vos_voc}, although the objects of the PASCAL VOC dataset undergo drastic variation in scale, position and appearance, our method improves upon the second-best results \cite{coseg_li19} by margins  1.1\% and 6\% in terms of $\mathcal{P}$ and $\mathcal{J}$ respectively. We also present some co-segmentation results of the proposed method in Figure \ref{pascal_sample}. It can be seen that our method can generate promising object segments under different types of intra-class variations, such as colors, sharps, views, scales and background clutters.
\begin{table*}[t!]
	\caption{Ablation study  on  DAVIS16 with different components used and  different numbers of D-features adopted. We also compare the performance  for different numbers of input images  on DAVIS16 and PASCAL VOC. }
	\centering
	\setlength{\tabcolsep}{1.0mm}{
		\begin{tabular}{l|ccccc}
			\toprule
			Method &Baseline &+DFM\&ATM &+ATM\&CRF &+multiple scales&+I.Prun.\\
			$\mathcal{J}$ mean (\%)&76.7         &79.5&80.4&81.1&83.4\\
			\hline
			\multicolumn{6}{c}{D-features }\\  \hline
			K &128&256 &512&1024     &2048\\
			$\mathcal{J}$ mean (\%)&78.3&79.0        &79.4&79.7&80.4\\
			\hline 
			\multicolumn{6}{c}{Input images (DAVIS16)}\\
			\hline
			$N^{in}$ & 1 &2    &4&8&10\\
			$\mathcal{J}$ mean (\%)&79.4        &80.1&80.4&80.4&80.4\\
			\hline 
			\multicolumn{6}{c}{Input images (PASCAL VOC)}\\  \hline
			$N^{in}$ & 1 &2    &4&8&10\\
			$Avg. \mathcal{J}$ (\%)&61.4         &63.5&65.0&65.4&65.4\\
			\bottomrule
		\end{tabular}
	}	
	\label{vos_ablation1}
\end{table*}
\subsection{Ablation Study}\label{exp_ablation}
To verify the effectiveness of the proposed method, we conduct ablation experiments on DAVIS16 and PASCAL VOC. As shown in Table \ref{vos_ablation1}, the detailed results are reported for different experimental setup. We adopt the DeepLabv3 as the baseline, which is trained on the static image dataset, and achieve 76.7\% in terms of $\mathcal{J}$. After adding the proposed DFM and ATM into the network, the performance increase to 79.5\%, which validates the usefulness of modeling the long-term dependency. We then adopt  CRF to optimize the attention map by considering the smoothness and consistency, which improves the performance by 0.9\%. Multiple-scale inference and instance pruning (I.Prun.) are also used by following \cite{vos_andiff}. At last, we obtain the highest score of 83.4\% in terms of the region similarity $\mathcal{J}$, which outperforms state-of-the-art methods. By adopting different numbers of D-features, we can see that better results can be obtained with more discriminative features used. We also evaluate the impact of the number of input images during inference, and we report performance with different values of $N^{in}$ on  DAVIS16 and PASCAL VOC datasets. For DAVIS16, we can see the performance increases by adding more input frames from 1 to 4 and then keep stable. It can be observed that with more input images, especially from 1  to 8, the performance raises accordingly on PASCAL VOC. When more images are considered, the performance does not change obviously. 
\subsection{Model Analysis}\label{exp_analysis}
In Table \ref{vos_analysis}, we report the comparison with state-of-the-art methods on the number of network parameters  and inference time on DAVIS16. We can observe that DFNet reduces the model complexity with fewer parameters compared with COSNet \cite{vos_cosnet}, AGNN \cite{vos_agnn} and AnDiff \cite{vos_andiff}. For the inference comparison, we run the public code of other methods and our code on the same machine with  NVIDIA GeForce GTX 1080 Ti. The inference time includes the image loading and pre-processing time. In can be seen that our method shows a faster speed than these methods.
\begin{table*}[t!]
	\caption{The number of model parameters and inference time comparison with state-of-the-art methods.}
	\centering
	\setlength{\tabcolsep}{1.0mm}{
		\begin{tabular}{l|cccc}
			\toprule
			Method &COSNet \cite{vos_cosnet} & AGNN \cite{vos_agnn} & AnDiff \cite{vos_andiff} & Ours\\ \hline
			\# Parmeter (M) &81.2 & 82.3 & 79.3 & \textbf{64.7} \\ \hline
			Inf. Time (s/image) &0.45 & 0.53 &0.35 & \textbf{0.28} \\	
			\bottomrule
		\end{tabular}
	}	
	\label{vos_analysis}
\end{table*}
\section{Conclusion}
To model the long-term dependency of video images, we propose a novel DFNet to capture the relations among video frames and infer the common foreground objects in this paper. It extracts the discriminative features from the input images, which can describe the feature distribution from a global view. An attention module is then adopted to mine the correlations between the input images.
The smoothness and consistency of the attention map are also considered, in which  the attention mechanism is  formulated as a classification problem and solved by CRF. The extensive experiments validate the effectiveness of the proposed method. In addition, we also apply the method to image object co-segmentation task. The quantitative evaluation of the challenging dataset PASCAL VOC demonstrates the advantage of DFNet. \\
\textbf{Acknowledgments.} This work is supported by Hong Kong RGC GRF 16206819, Hong Kong RGC GRF 16203518 and Hong
Kong T22-603/15N.

\clearpage
%
%
\bibliographystyle{splncs04}
\bibliography{vosnet}

\end{document}